\documentclass[review]{elsarticle}

\usepackage{lineno,hyperref}
\usepackage{framed,multirow}
\usepackage{booktabs}
\usepackage{amssymb}
\usepackage{latexsym}
\usepackage{algorithm, algorithmic}
\usepackage{url}
\usepackage{xcolor}
\usepackage{amsmath}
\usepackage{pifont}     
\usepackage{threeparttable}
\usepackage{bbding}   
\usepackage{fontawesome}  
\usepackage[shortlabels]{enumitem}
\modulolinenumbers[5]

\journal{Journal of \LaTeX\ Templates}

\begin{document}

\begin{frontmatter}

\title{The Binary Quantized Neural Network for Dense Prediction via Specially Designed Upsampling and Attention}

\author[1]{Xingyu Ding}
\ead{dingxingyu21@mails.ucas.ac.cn}

\author[1]{Lianlei Shan\corref{mycorrespondingauthor}}
\ead{shanlianlei18@mails.ucas.ac.cn}

\author[1]{Guiqin Zhao}
\ead{zhaoguiqin20@mails.ucas.ac.cn}

\author[1]{Meiqi Wu}
\ead{wumeiqi18@mails.ucas.ac.cn}

\author[2]{Wenzhang Zhou}
\ead{20230196@njupt.edu.cn}

\author[3]{Wei Li}
\ead{leesoon@bupt.edu.cn}

\address[1]{University of Chinese Academy of Sciences, Beijing, China}
\address[2]{Nanjing University of Posts and Telecommunications, China}
\address[3]{Beijing University of Posts and Telecommunications, China}
\cortext[cor1]{Corresponding author}

\begin{abstract}
Deep learning-based information processing consumes long time and requires huge computing resources, especially for dense prediction tasks which require an output for each pixel, like semantic segmentation and salient object detection.
There are mainly two challenges for quantization of dense prediction tasks. Firstly, directly applying the upsampling operation that dense prediction tasks require is extremely crude and causes unacceptable accuracy reduction. Secondly, the complex structure of dense prediction networks means it is difficult to maintain a fast speed as well as a high accuracy when performing quantization. 
In this paper, we propose an effective upsampling method and an efficient attention computation strategy to transfer the success of the binary neural networks (BNN) from single prediction tasks to dense prediction tasks. 
Firstly, we design a simple and robust multi-branch parallel upsampling structure to achieve the high accuracy. Then we further optimize the attention method which plays an important role in segmentation but has huge computation complexity. 
Our attention method can reduce the computational complexity by a factor of one hundred times but retain the original effect.
Experiments on Cityscapes, KITTI road, and ECSSD fully show the effectiveness of our work.

\end{abstract}

\end{frontmatter}


\section{Introduction}
\begin{figure}[!h]
\centering
\includegraphics[scale=0.08]{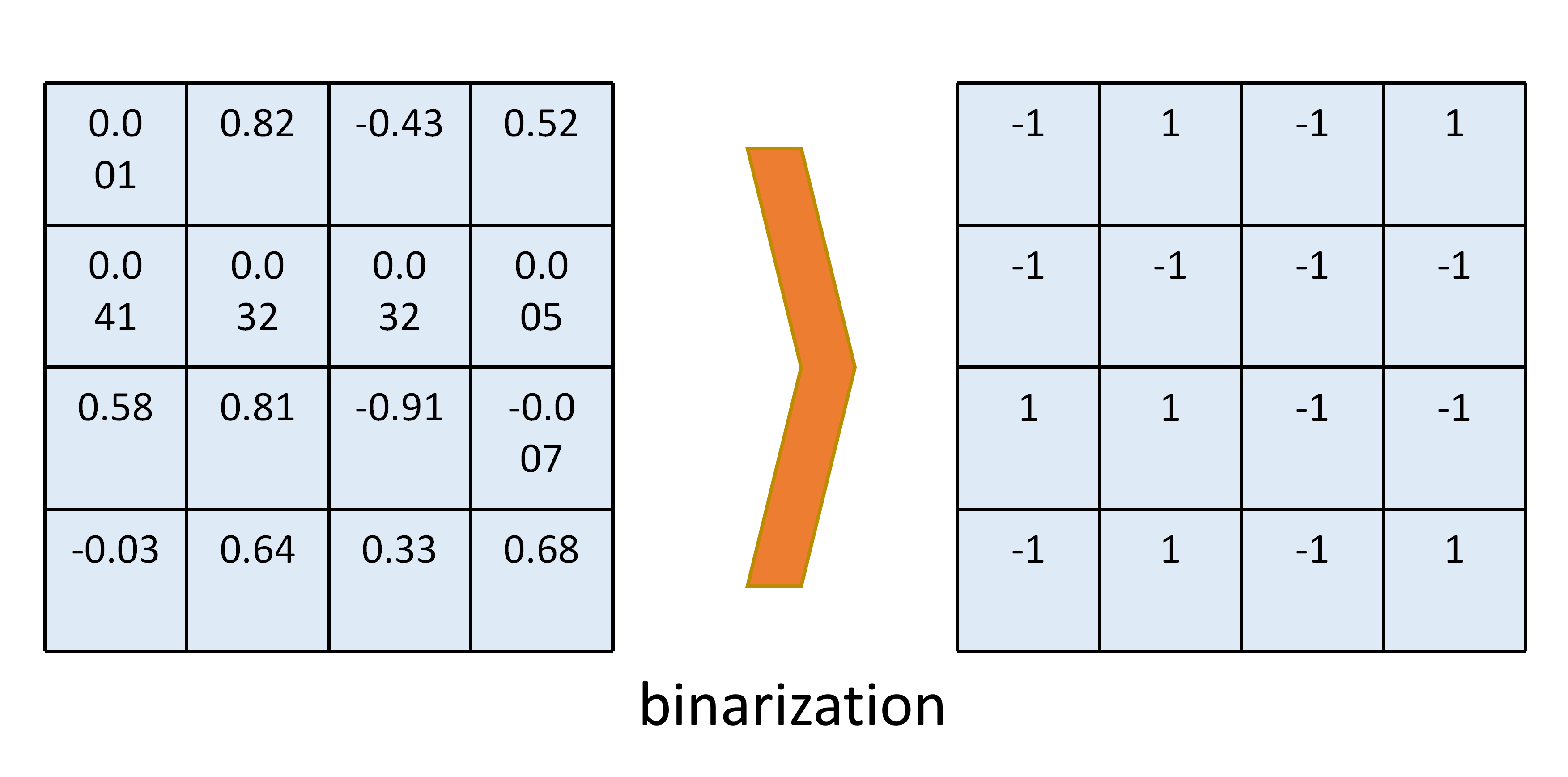}
\caption{
The procedure of binarization.
On the left is the feature map or the convolution kernel, where each number is 16-bit with full precision.
After binarization, as shown on the right, the value is the binary with 1-bit.
}
\label{binarization_intro}
\end{figure}

Quantization is an important research topic in model compression and acceleration and can decrease dozens of folds in both occupied storage and computational complexity.
Quantization, especially binarization, achieves tremendous improvements in classification recently, which provides a good foundation for the application of dense prediction tasks. 
As is shown in Fig. \ref{binarization_intro}, binarization changes the values expressed with full precision in the feature maps or the convolution kernels to binary values.

X-NOR \citep{xnor}, and Bi-Real Net \citep{liu2018bi} enable binary networks to match the full precision networks at the component level.
However, due to the high requirements of dense prediction tasks, it is not enough to only perform binarization in component level, and the network structure also needs to be adjusted. 
Transfering the network structures of full-precision directly to BNN makes the accuracy drop a lot.
 
Previous quantizations of dense prediction generally use four-valued or eight-valued quantization to reduce the drop of accuracy \citep{unetfix}. These methods are commonly used in tasks with fewer feature changes like medical or remote sensing images. However, compared with binary quantization, these multi-valued quantization methods bring much less benefits in acceleration and compression. Especially for acceleration, binary networks can replace arithmetical operations with logical operations in hardware to increase the speed by dozens of folds, which is the advantage that multi-valued quantizations do not contain.

Although binarization contains huge advantages in storage and computation, the expressive ability of the network is limited after binarization, which is fatal to dense prediction tasks. A lot of works improve this from the structure level.
WRPN \citep{wrpn} and \cite{shen2019searching} improve the accuracy of BNN by increasing the number of channels after each convolution layer.
ABC-Net \citep{lin2017towards} and groupNet \citep{group,group2} adopt group structure.
Although this approach increases computational complexity and loses part of the speed advantage, this sacrifice is necessary for dense prediction tasks. We follow this structure and propose the upsampling and attention computation of dense prediction tasks based on this.

GroupNet not only discusses the effect in image classification but also discusses the effect in dense prediction tasks like segmentation.
GroupNet uses multiple parallel binary operations to approximate the original full-precision operation, which not only retains the powerful compression and acceleration advantages of BNN but also retains the original information to the maximum extent.
However, in groupNet, only the dilated convolution as decoder head is discussed, but the upsampling as decoder head are not discussed.
Moreover, dilated convolution is not suitable for semantic segmentation of binary networks, when the rate=\{12,24,36\}, Atrous Spatial Pyramid Pooling (ASPP) \cite{deeplabv3} needs a lot of padding, and filling with $-1$ or $1$ is both prone to produce errors.
Encoder-decoder structure with upsampling are mostly used in dense prediction tasks \citep{unet,deeplabv3}, so upsampling is unavoidable if BNN is to be developed and applied in dense prediction.
Our contribution is to maintain the upsampling of semantic segmentation while keeping the huge acceleration advantage of BNN.
Compared with a full-precision or multi-valued quantized network, the upsampling of BNN, whether bilinear interpolation or nearest-neighbor interpolation, is of little significance and thus leads to poor performance.
Therefore, inspired by groupNet, we design a multi-branch parallel upsampling method to solve the problem.
The main idea is to send the original feature maps to multiple parallel branches for upsampling and then merge them together, which can approximate the full precision result without reducing the inference speed.

Besides upsampling, attention also has essential significance in dense prediction tasks like semantic segmentation and image defogging, but its huge computational complexity hinders its wide application.
And we also find that the attention mechanism can be easily migrated from the full precision network to BNN with this structure.
In addition, we analyze the differences between our design with Inception \citep{inception} and methods of increasing the number of channels and demonstrate the principle for the effectiveness of our proposed structure from the perspective of ensemble learning.

To sum up, our contributions are summarized as follows,
\begin{itemize}
\item[$\bullet$]We design a multi-branch parallel upsampling method to make the upsampling in BNN obtain competitive results compared with the full precision network without slowing inference speed.
\item[$\bullet$]The same parallel strategy is also used to adapt the attention mechanism from the full-precision network to the binary network.
Our method can be applied to the original attention calculation method and can be used for both spatial attention and channel attention.
\item[$\bullet$]We explain the effectiveness of the proposed structure from the perspective of the ensemble learning and analyze the factors that limit the accuracy of the BNN.
\item[$\bullet$]Extensive experiments are conducted on Cityscapes, KITTI road, and ECSSD, and results are competitive in segmentation accuracy, compression, and acceleration.
\end{itemize}

\section{Related Work}
\subsection{Quantization of dense prediction networks}

Dense prediction includes semantic segmentation \cite{shan2021decouple,shan2021class,shan2021densenet,shan2022mbnet,shan2021uhrsnet,shan2023boosting,shan2023incremental,shan2023data,shan2022class,wu2023continual} and others \cite{zhao2023explore,zhao2023flowtext,zhao2023generative,zhao2024controlcap}. 
The quantitative research on dense predictive networks mainly focuses on the quantization of semantic segmentation networks.
Direct binarization results in poor performance, and the previous approaches make one compromise: using multi-valued quantization.
\cite{unetfix} achieves an eight-fold reduction in memory requirements with losing very little precision via four bits of weight and six bits of activation.
\cite{3dq} propose a 3-bit quantization method, based on FCN \citep{fcn} and U-Net \citep{unet}, which achieves good results in human brain segmentation.
\cite{yang2020training} uses 8-bit quantization and removes square and root operations in batch normalization to give the network the ability to inference on fixed-point devices, which achieves competitive results.
groupNet and Binary Dad-Net \citep{frickenstein2020binary} firstly introduces BNN to semantic segmentation, but they miss the analysis of upsampling.
They use full-precision upsampling on the last layer based on FCN and uses dilated convolution based on deeplabv3 \citep{deeplabv3} to avoid upsampling.
However, the mainstream framework of semantic segmentation networks, such as U-Net \citep{unet}, PSPNet \citep{pspnet} and others \citep{segmentation_1,segmentation_2}, all require upsampling. Therefore, in order to achieve great further development of BNN in semantic segmentation, upsampling is impossible to bypass.
After a careful search, for dense peidiction like salient object detection or image deblurring, there is no accelerated work through quantification, and the previous are to design lightweight real-time network structures.
Because network quantization has unique advantages in practical deployment that lightweight networks do not have, it is meaningful to quantify saliency detection networks and image deblurring networks.

In summary, compared with previous works, our contribution is to eliminate the accuracy drop caused by upsampling in binary networks while maintaining the excellent acceleration and compression characteristics of binary networks.

\subsection{Quantization of Attention}
Attention plays a vital role in semantic segmentation and salient object detection and gains tremendous improvement after non-local \citep{non-local}.
Channel Attention is proposed by SENet \citep{senet}, and SKNet \citep{sknet}.
DANet \citep{danet} aggregates channel and spatial attention into one network and gets best results in panoptic segmentation.
\cite{tao2020hierarchical, attention_1,attention_2}, propose hierarchical multi-scale attention, which is the current SOTA result.
\cite{aaa} uses Pyramid Feature Attention Network for salient target detection.
BANet \citep{banet} uses the bilateral attention model to achieve rapid and efficient salient target detection.
Grid-Dehaze-Net \citep{grid} proposes a multi-scale dehazing network based on an attention mechanism.
PANet \citep{panet} proposes an image restoration network based on a pyramid attention network, which can be used for denoising, deleting mosaics, and super-resolution.

It can be seen that the attention mechanism plays an important role in dense prediction tasks, so it is necessary to transfer it from full precision to a binary network.
And the most deadly problem of this transferring is binary values do not satisfy the expressive hierarchy of attention.
Since the convolution parameters and input features are all binary, the attention level has changed from an infinite number of full-precision networks to only two levels. This low degree of differentiation makes attention less effective.
This problem can also be solved by the multi-branch parallel structure.
To the best of our knows, we are the first to employ attention in BNN, and other techniques in full-precision semantic segmentation such as edge monitoring can also be migrated to binary networks in a similar manner.

\section{Proposed Method}
\begin{figure}[!h]
\centering
\includegraphics[scale=0.06]{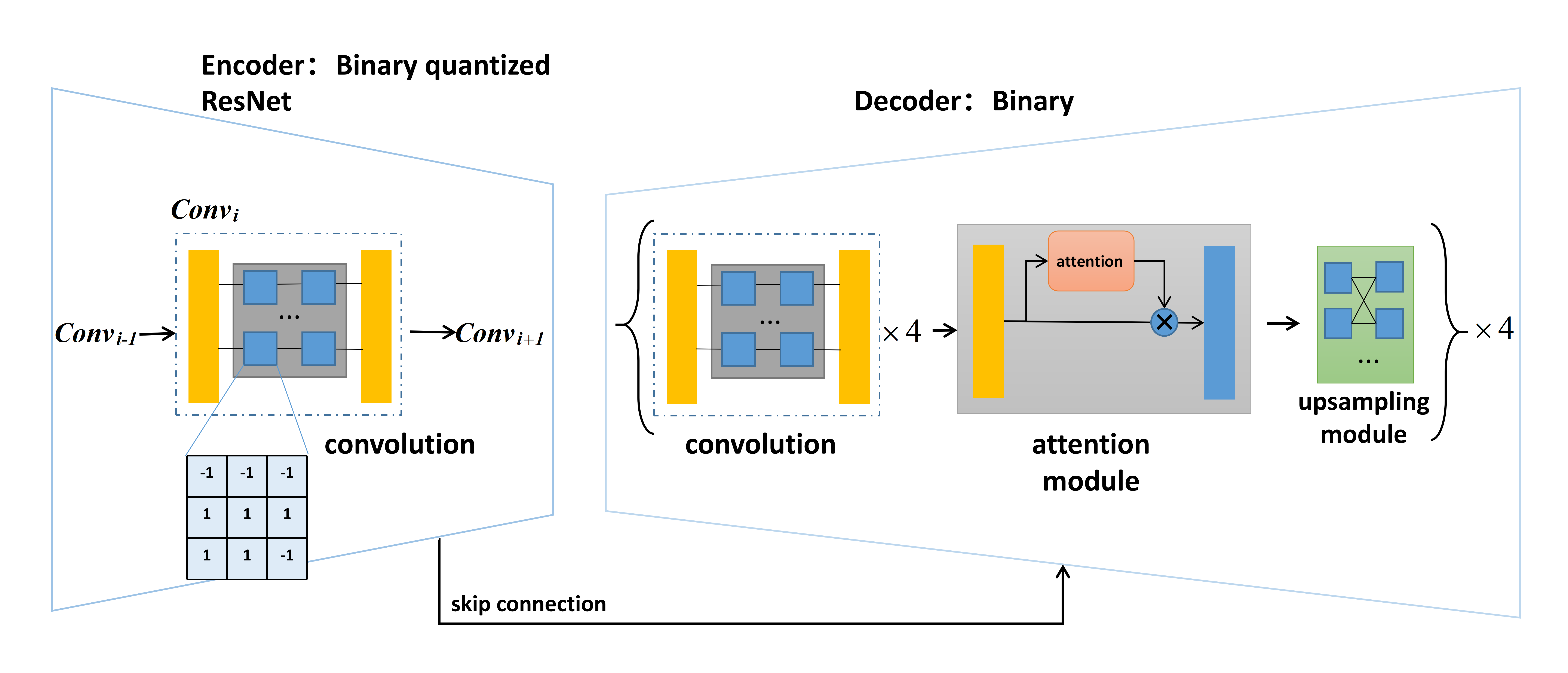}
\caption{Overview of the entire network structure.
The whole network structure is consistent with U-Net \citep{unet}.
The encoding part is ResNet, and the decoding part adopts the structure of FPN \citep{fpn}. 
In the decoding part, there are four large modules, each of which is composed of four convolution modules, one attention module, and one upsampling module.
All operations are binarized.
}
\label{overview_pic}
\end{figure}

\subsection{Overview}

The overview of the entire network structure is shown in Fig. \ref{overview_pic}. We adopted the network structure similar to U-Net because there are many up-sampling operations in U-Net, which is easier to show the significance of our work.
All operations in the network, including convolution, upsampling, and so on, are all binary.
Our two major improvements are upsampling module and attention module, shown in the orange and gray parts in the figure.
In section \ref{quantization_sec}, we introduce the used quantization method.
We introduce binary upsampling in section \ref{upsample_section} and binary attention in section \ref{attention_section}.

\subsection{Quantization Method}
\label{quantization_sec}
Network binarization is the binarization of convolution parameters and input features, which corresponds to the binarization of weights and activations.
We implement the same quantization functions as groupNet, and more details can be found in \cite{group}.
For a convolutional layer, we define the input $\mathbf{x} \in \mathbb{R}^{c_{in} \times w_{in} \times h_{in}},$ weight filter $\mathbf{w} \in \mathbb{R}^{c \times w \times h}$ and the output $\mathbf{y} \in \mathbb{R}^{\text {cout} \times w_{\text {out}} \times h_{\text {out}}},$ respectively.

. 
\textbf{Binarization of weights}: Following X-NOR, the floating-point weight $w$ is approximate by a binary weight filter $b^{w}$ and a scaling factor $\alpha \in \mathbb{R}^{+}$ to ensure $w \approx \alpha b^{w}$, where $b^{w}$ is the sign of $w$ and $\alpha$ calculates the mean of absolute values of $w$.
$\operatorname{sign}(\cdot)$ is non-differentiable and we adopt the straight-through estimator (STE) \cite{ste} to approximate the gradient calculation. Formally, the forward and backward processes can be given as follows:

\begin{equation}
\centering
\begin{array}{l}
\text { Forward }: \mathbf{b}^{w}=\operatorname{sign}(\mathbf{w}) \\
\text { Backward : } \frac{\partial \ell}{\partial \mathbf{w}}=\frac{\partial \ell}{\partial \mathbf{b}^{w}} \cdot \frac{\partial \mathbf{b}^{w}}{\partial \mathbf{w}} \approx \frac{\partial \ell}{\partial \mathbf{b}^{w}}
\end{array}
\label{weight_1}
\end{equation}

where $\ell$ is the loss.

\textbf{Binarization of activations}: For activation binarization, the piecewise polynomial function to approximate the sign function as in Bi-real-Net is most suitable. The forward and backward can be written as:

\begin{equation}
\centering
\begin{array}{l}
\text { Forward }: b^{a}=\operatorname{sign}(\mathrm{x}) \\
\text { Backward: } \frac{\partial \ell}{\partial x}=\frac{\partial \ell}{\partial b^{a}} \cdot \frac{\partial b^{a}}{\partial x} \\
\text { where } \frac{\partial b^{a}}{\partial x}=\left\{\begin{array}{l}
2+2 x:-1 \leq x<0 \\
2-2 x: 0 \leq x<1 \\
0: \text { otherwise }
\end{array}\right.
\end{array}
\end{equation}

\subsection{Upsampling of The Binary Neural Network}
\label{upsample_section}

\begin{figure}[H]
\centering
\includegraphics[scale=0.09]{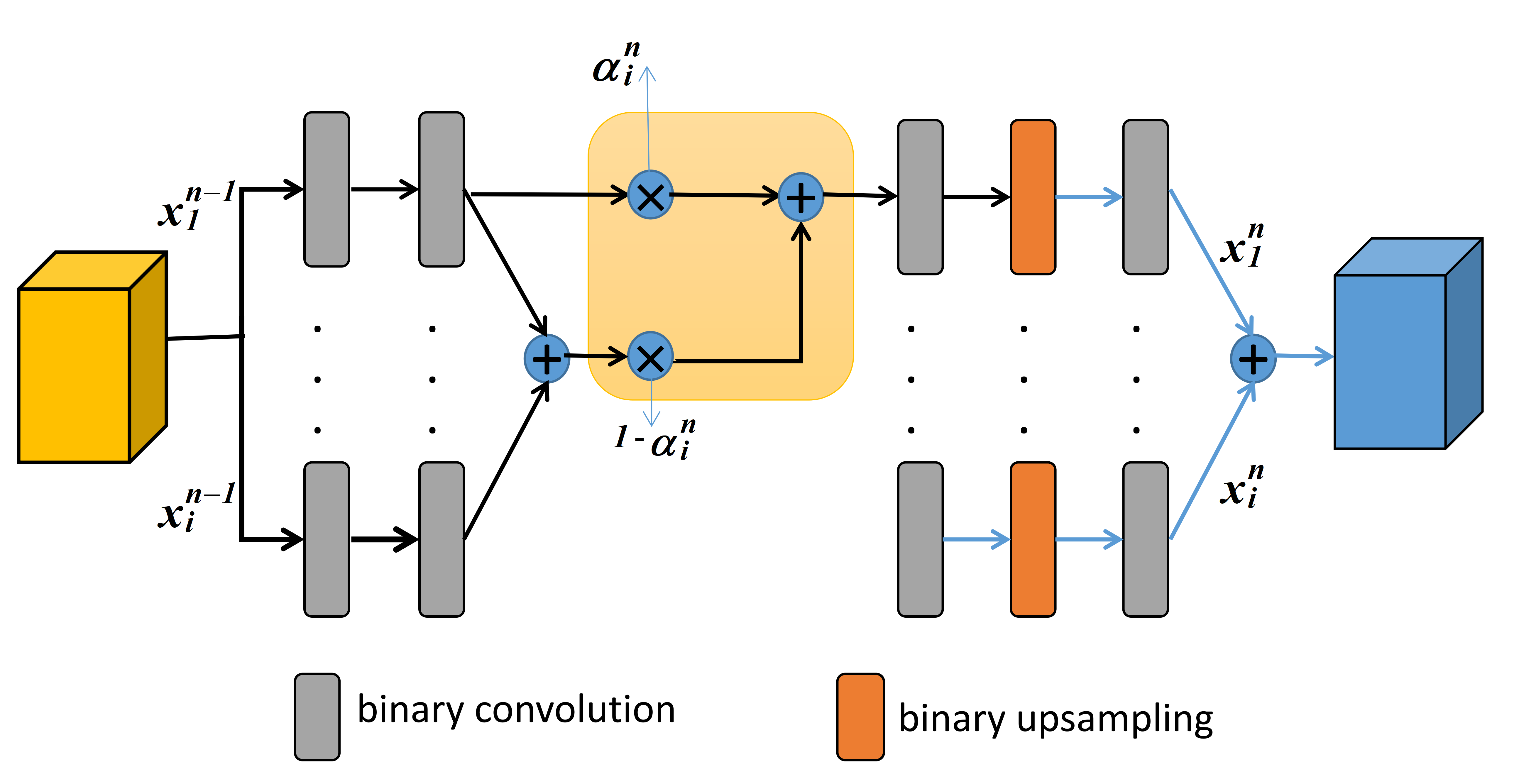}
\caption{Overview of the proposed upsampling module.}
\label{upsample_pic}
\end{figure}

The role of our proposed upsampling module of BNN is consistent with the role in full-precision networks.
We first introduce the structure of upsampling, then analyze the computational complexity to both full-precision and fixed-point quantization networks.

\subsubsection{Structure}

Our proposed upsampling method is shown in Fig. \ref{upsample_pic}, which is the multi-branch parallel structure. 
As shown in Fig. \ref{upsample_pic}, the features that need to be upsampled are copied several times and then be put into different branches, respectively. These branches have the same structure, but the parameters are not shared.
The location of the orange part in the figure is behind the second convolution block, and the purpose of which is to obtain the input of the next convolution via the output of the previous convolution.

As shown in orange part, we define the input of the $n$-th convolution at the $i$-th branch from the ($n$-1)-th convolution as Eq. \ref{eq_3} and \ref{parameter_form}:

\begin{equation}
\label{parameter_form}
\mathbf{x}_{i}^{n}={\alpha}_{i} \odot B_{i}^{n-1}\left(\mathbf{x}_{i}^{n-1}\right)+
\left(1-{\alpha}_{i}\right) \odot \sum_{j=1}^{K} B_{j}^{n-1}\left(\mathbf{x}_{j}^{n-1}\right)
\end{equation}
\begin{equation}
{\alpha}_{i} =\operatorname{sigmoid}\left(\theta_{i}\right)
\label{eq_3}
\end{equation}
where $\theta \in \mathbb{R}^{K}$ is a parameter vector that needs to be learned, $\alpha$ is a gate scalar. $K$ is the number of branches. $B$ is the binary convolution, which is shown in Eq. \ref{binary_conv_form}. The binary upsampling is the nearest neighbor upsampling.
\begin{equation}
B_{i}(\mathbf{x})= \lambda_{i} \odot (\mathbf{b}_{i}^{w} \oplus \operatorname{sign}(\mathbf{x}))
\label{binary_conv_form}
\end{equation}
where $\oplus$ is bitwise operations $\operatorname{xnor}(\cdot)$ and $\operatorname{popcount}(\cdot)$. $\odot$ is the Hadamard product.

Finally, the final feature map $\mathbf{x}^{n}$ is obtained through Eq. \ref{eq5}, where $\lambda$ is also a learned scalar, and the calculation method is the same with $\alpha$.
\begin{equation}
\label{eq5}
\mathbf{x}^{n}=\sum_{i=1}^{K} \lambda_{i} \odot \mathbf{x}_{i}^{n}
\end{equation}

Our structure is borrowed from the soft-gate in groupNet, but the soft-gate is between each block, and ours is inside one block.
The purpose of groupNet is not to lose too much information as features travel between blocks, but ours is to prevent some branches with particularly large errors.
Correcting this branch with all the rest information rather than only this branch will inevitably bring more input information to each branch, and also make all the branches not appear particularly big errors, so as to greatly enhance the robustness of the network, which is the most needed for the binarization network.
Therefore, the structure is similar, but the purpose is completely different.

The proposed structure is similar to ResNext \cite{resnext} and Inception \cite{inception}, but actually there are essential differences in both motivation and specific operations.
ResNext and Inception are designed to reduce parameters and computation complexity, both of which have operations to reduce the number of channels in each branch, and the difference of the two is whether each branch is same with others.
However, in the semantic segmentation of BNN, our motivation is to increase the expressive ability of network and reduce the errors caused by inaccurate fillings during upsampling.
Different purposes lead to different operations, and the most obvious difference is that we do not reduce the number of channels in each branch.

\subsubsection{Complexity analysis}
We compare our method with full-precision upsampling and fixed-point quantized upsampling.
During inference, the homogeneous $K$ branches can be parallelizable, and thus the structure is friendly to hardware. 
Specifically, the bitwise X-NOR operation and bit-counting can be performed in a parallel of $64$ by the current generation of CPUs \citep{liu2018bi}, which is a huge advantage that multi-valued quantization does not have. 
This will bring significant gain in speed-up of the inference compared with not only full-precision but also multi-valued quantization. 

\textbf{Compared with full-precision network:}
Complexity of convolution, addition, and upsampling shown as Eq. \ref{complexity_form}.
 \begin{equation}
 \label{complexity_form}
\left\{
\begin{array}{l}
C_{conv}=c_{in} \cdot c_{out} \cdot w \cdot h \cdot w_{in} \cdot h_{in}\\
C_{add}=c_{out} \cdot w_{out} \cdot h_{out}\\
C_{up}=c_{in} \cdot w_{in} \cdot h_{in}\\
\end{array}
\right.
\end{equation}
where the meaning of the parameter is the same as in Eq. \ref{weight_1}.

We just need to calculate $K$ binary convolutions and $K$ full-precision additions. As a result, the speed-up ratio $\sigma$ for a convolutional layer can be calculated as,\\

\begin{equation}
\begin{aligned}
&\sigma =\frac{C_{up}+C_{conv}}
{ K\cdot C_{up}+\frac{1}{64}\cdot K \cdot C_{conv}+K\cdot C_{add}}\\
&=\frac{64}{K} \cdot \frac{ C_{conv}+\cdot C_{up}}{C_{conv}+64\cdot  C_{up}+64\cdot  C_{add}}
\end{aligned}
\end{equation}

We take the upsampling that makes the image twice its original size as an example.
If we set $c_{in}=c_{out}=$256$, w \times h=3 \times 3, w_{in}=h_{in}=40,w_{out}=h_{out}=80$, $K=5$, then it can reach $11.25 \times$ speedup. But in practice, each branch can be implemented in parallel. And the actual speedup ratio is also influenced by the process of memory read and thread communication.

\textbf{Compared with fixed-point quantization network:}
Our $K$ multi-branch parallel structure is different from the $K$ bit fixed-point approaches.
The inner product between fixed-point weights and activations can be computed by bitwise operations.
The weight vector $\mathbf{w} \in \mathbb{R}^{M}$ can be encoded by a vector $\mathbf{b}_{i}^{w} \in\{-1,1\}^{M}, i=1, \ldots, K$.
Activations is also quantized to $K$-bit. So, similarly, the activations $\mathbf{x}$ can be encoded by $\mathbf{b}_{j}^{a} \in\{-1,1\}^{M}, j=1, \ldots, K$. Thus, the convolution can be written as
\begin{equation}
Q_{K}\left(\mathbf{w}^{T}\right) Q_{K}(\mathbf{x})=\sum_{i=0}^{K-1} \sum_{j=0}^{K-1} 2^{i+j}\left(\mathbf{b}_{i}^{w} \oplus \mathbf{b}_{j}^{a}\right)
\end{equation}
where $Q_{K}(\cdot)$ is any uniform quantization function.
It calculates and sums over $K^{2}$ times $\operatorname{xnor}(\cdot)$ and $popcount (\cdot)$. The complexity is about $O\left(K^{2}\right)$, and the output range for a single output is $\left[-\left(2^{K}-1\right)^{2} M,\left(2^{K}-1\right)^{2} M\right]$.
From Eq. \ref{binary_conv_form}, our computational complexity is $O(K)$, and the value range for each element after summation is $[-KM, KM]$, which is much less than that in fixed-point methods.

So, similar to groupNet, our approach has huge advantages in storage and acceleration compared with fixed-point methods.

\subsubsection{Visual presentation}
\begin{figure}[!h]
\centering
\includegraphics[scale=0.08]{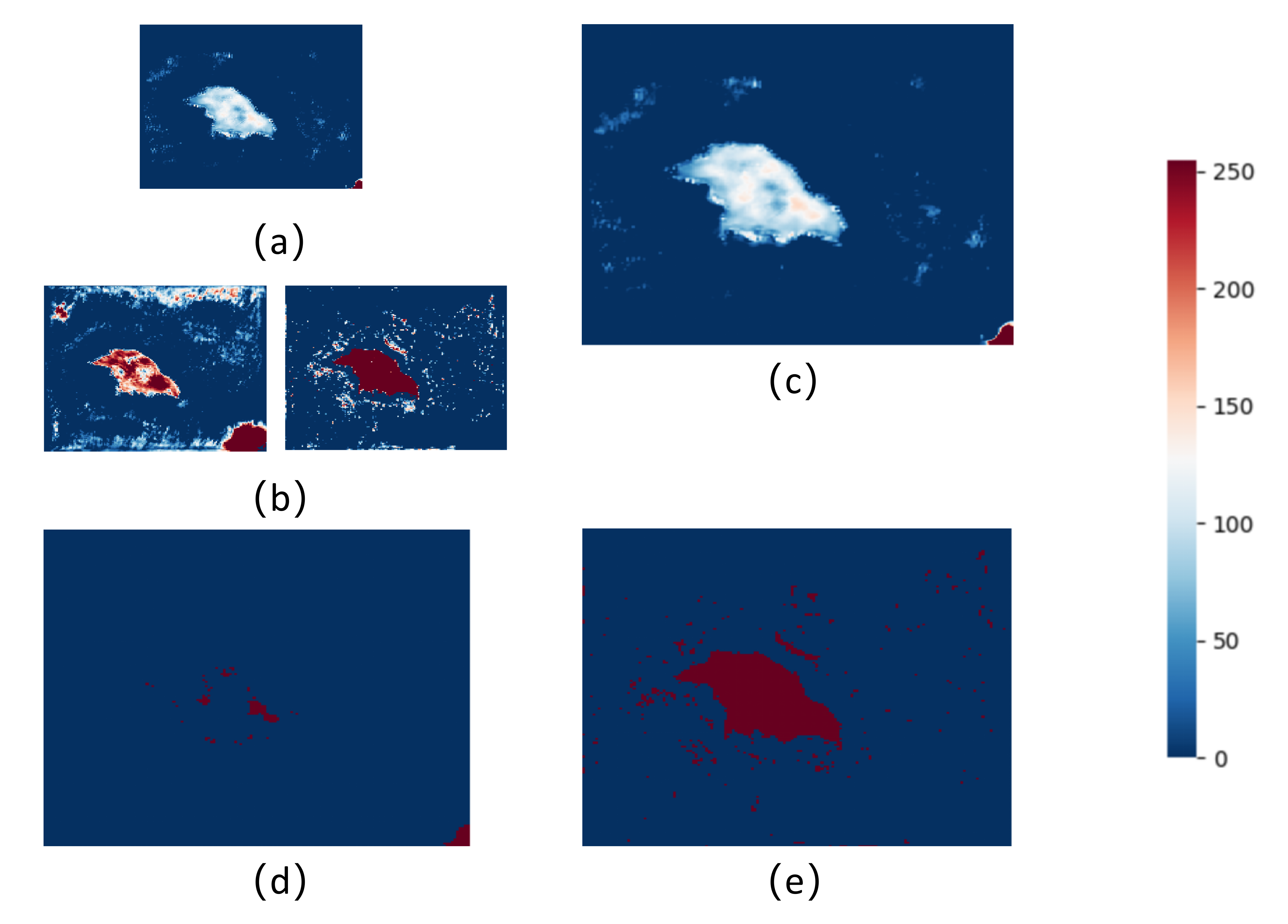}
\caption{(a) is the feature map requiring up-sampling, (b) are the feature maps providing auxiliary information in our method, (c) is the result of full precision up-sampling, (d) is the result of binary up-sampling, and (e) is the result of our binary up-sampling.
It can be seen that the results of our method are more similar to those of the full-precision method.}
\label{upsample_vis}
\end{figure}
Fig. \ref{upsample_vis} visually demonstrates the effect of our method.
It can be seen that if the previous method is used for upsampling with only the information of the one channel, the result shown in (d) will be obtained, and a large amount of information will be lost.
On the contrary, through our method, the result shown in (e) can be obtained by using the information of surrounding channels such as (b) during the upsampling.
(c) is the result of full precision upsampling, and it can be seen that the result obtained by our upsampling method is very similar to the result of full precision, which proves the effectiveness of our method.

\subsection{Binary Attention}

\begin{figure}[!h]
\centering
\includegraphics[scale=0.08]{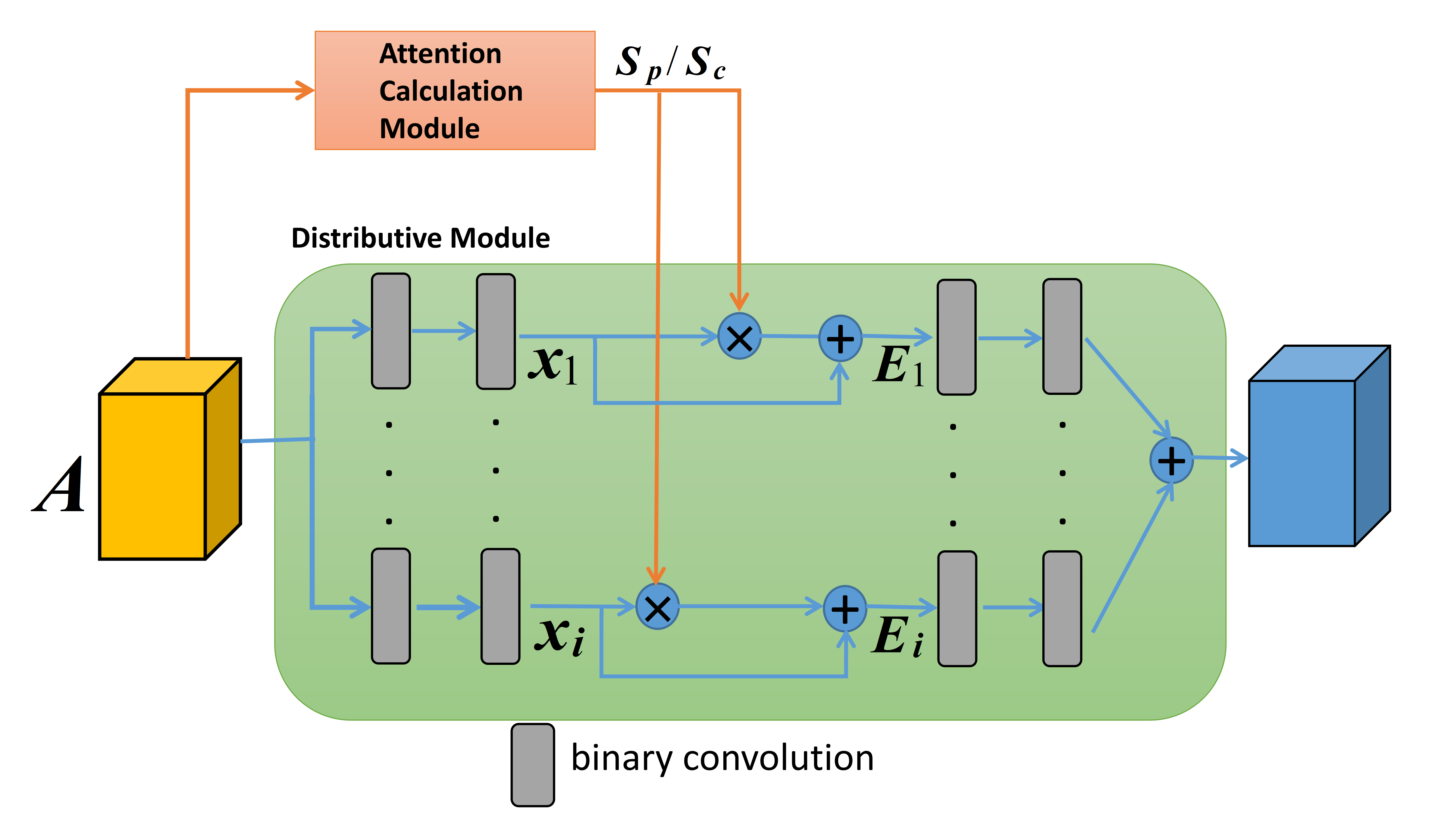}
\caption{Overview of the proposed attention module.}
\label{attention_pic_1}
\end{figure}

\label{attention_section}
\subsubsection{Structure}
The binary attention is shown in Fig. \ref{attention_pic_1}, which consists of two parts: attention calculation module and attention distributive module.
The attention calculation module can be any attention computing module proposed in full-precision semantic segmentation networks.
The overall process is calculating the attention matrix from the original feature firstly and then multiplying feature maps of each branch with the calculated attention.

We feed feature $\mathbf{A}$ into a convolution layer to generate a new feature map $\mathbf{x} \in \mathbb{R}^{C \times H \times W}$ and reshape it to $\mathbb{R}^{C \times M}$. Then we perform a matrix multiplication between $\mathbf{x}$ and the transpose of $\mathbf{S}_{p}$ and reshape the result to $\mathbb{R}^{C \times H \times W}$. Finally, we obtain the final output $\mathbf{E} \in \mathbb{R}^{C \times H \times W}$ as follows:

\begin{equation}
\mathbf{E}_{i}=\beta \odot (\mathbf{S}_{p} \mathbf{x}_{i} +\mathbf{S}_{c} \mathbf{x}_{i})+\mathbf{x}_{i}
\label{eq9}
\end{equation}

where $\beta$ is initialized as 0 and gradually learns to assign more weight \citep{29}. 
$\mathbf{S}_{p}$ and $\mathbf{S}_{c}$ are the spatial and channel attention map obtained by the attention calculation module, which is described below.

\subsubsection{Complexity analysis}

The complexity of channel and positional attention is calculated in a similar way, so we only show the calculation process of channel attention.

The computational complexity and allocation complexity of channel attention are respectively shown as follows,
\begin{equation}
\left\{
\begin{array}{l}
C_{comp}=c_{in} \cdot \left (h_{in} \cdot  w_{in} \right )\cdot \left (w_{in} \cdot h_{in}\right )\\
C_{allo}=c_{in} \cdot c_{in} \cdot w_{in} \cdot h_{in}\\

\end{array}
\right.
\end{equation}
Where $C_{comp}$ and $C_{allo}$ represent the complexity of computation and allocation respectively, and the meaning of other symbols are the same as that in Eq. \ref{complexity_form}.

The speed-up ratio $\sigma$ compared with full precision:

\begin{equation}
\begin{aligned}
\sigma
& =\frac{C_{comp}+C_{allo}}{
\frac{1}{64}\cdot \left( C_{comp}+K \cdot C_{allo} \right)+K\cdot C_{add}}\\
&=\frac{64}{K} \cdot \frac{ C_{comp}+ C_{allo}}{\frac{1}{K}C_{comp}+C_{allo}+64 \cdot C_{add}}
\end{aligned}
\end{equation}

If use the same parameters as in section \ref{upsample_section}, then it can reach $137.90 \times$ speedup. 
The computational complexity of attention has always been a problem in semantic segmentation, and there are many efforts to reduce the computational complexity of attention, but such a problem does not exist in BNNs.
There is a lot of multiplication in the calculation of attention, and the acceleration of multiplication by BNN is most significant.

\subsubsection{Analysis of the effectiveness of attention binarization}
\begin{figure}[h]
\centering
\includegraphics[scale=0.06]{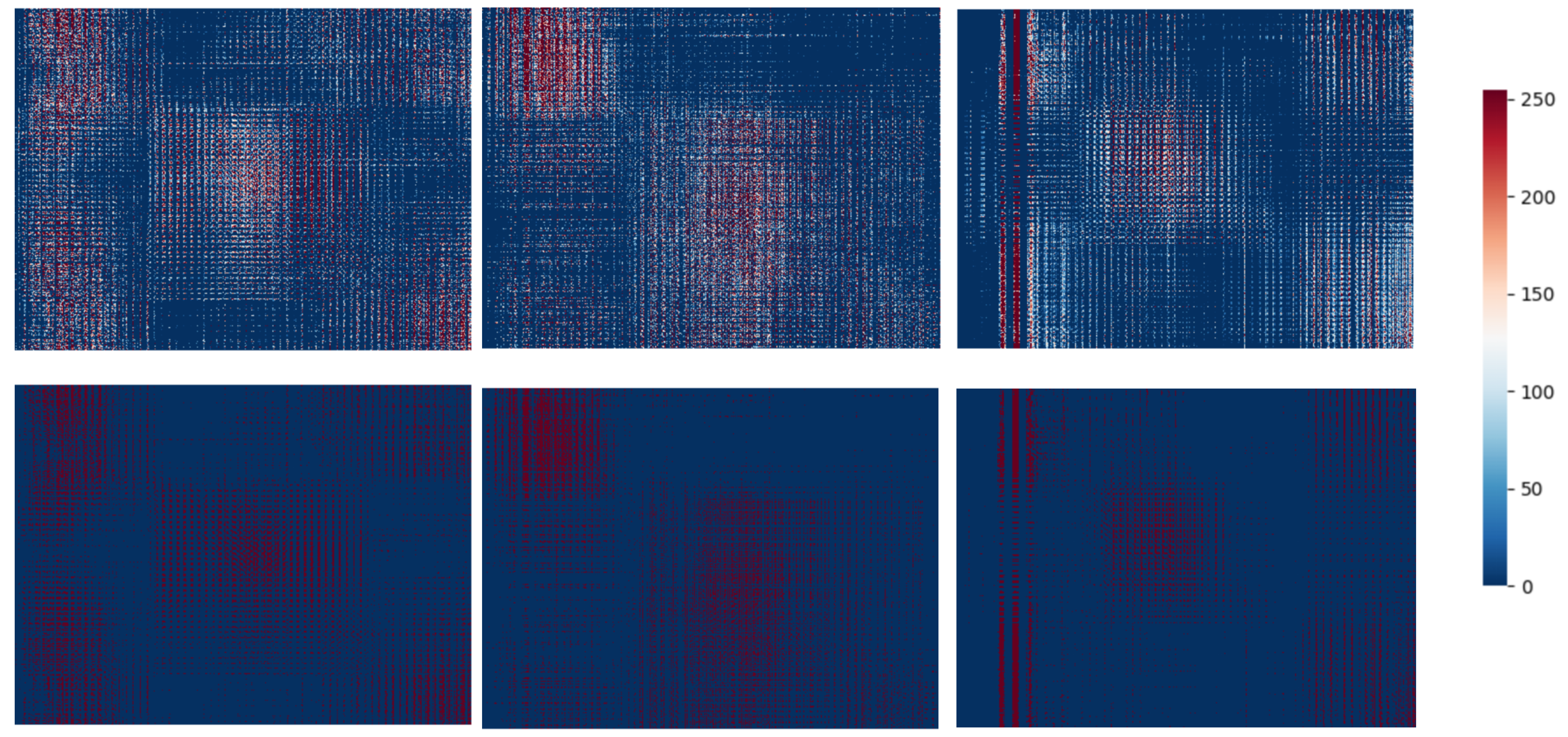}
\caption{Comparison of full precision attention maps and binary attention maps. The above is the full precision attention map, and the below is the binary attention map.}
\label{attention_pic}
\end{figure}
The lousy performance of binary attention is due to the property of binary multiplication and the weak ability of distinction with only two values.
For the property of the binary multiplication, one value of the original feature is -1, and the corresponding value of the attention map is -1, so the result is 1. Meanwhile, if the original feature and the attention map are 1, then the result is also 1. This causes a great deal of confusion but can be easily avoided by using the ReLU \citep{relu} as the activation function.
After changing the activation, the value of the attention map is basically positive.
For the weak ability of distinction with only two values, because binarization only has two degrees, the correlation and non-correlation between pixels cannot be distinguished clearly, which is the main reason for the weak ability of binary attention.
The weakening of expression seems inevitable, but the computational complexity of binary attention is small to negligible, so it can be calculated on larger feature maps, which alleviates the problem of poor expressive ability to some extent.
Besides, the multi-branch structure makes its robustness greatly improved.
If the number of branches is $5$, the final result will be wrong in cases three or more branches are wrong, which is such a low probability.
Due to this, computing once and then distributing is similar to computing on each branch individually ineffectiveness and robustness but is more efficient.

As shown in Fig. \ref{attention_pic}, compared with the full precision attention map, the binary attention map is far inferior in the ability of degree hierarchy.
However, under the condition of guaranteed binary, the position accuracy of attention is retained to the maximum extent, which ensures that the binary attention map can play a corresponding effect.

\section{Experiments}
\subsection{Implementation Details}
The whole code is based on Pytorch Encoding \citep{pytorch-encoding}.
We first train the binary backbone network on ImageNet \citep{imagenet} and then fine-tune it. The pretrained binary ResNet \citep{resnet} is from groupNet.
As in \cite{3,55,57,xnor,group}, except the first and the last layer, we quantize the weights and activations of all convolutional layers.
We use random horizontal flipping, cropping, and image color augmentation (gamma shift, brightness shift) to augment.
We use simple single crop testing for normal evaluation. We first train the full-precision model as initialization and then fine-tune the binary counterpart. We use Adam \citep{adam} for optimization. 


The training data generator (TDG) \citep{self25} automatically generates annotations for the task of driveable area segmentation, and we use this for our training.
\subsubsection{Datasets}
\textbf{CityScapes:}The CityScapes dataset \citep{cityscapes} consists of 2975 training images, 500 validation images, and 1525 test images, including their corresponding ground truth labels.
However, during training, the original images are sampled down to a size of 1024 $\times$ 512.
Categories of road and parking areas marked by humans are considered drivable areas. The remaining categories are assigned to non-drivable areas.

\textbf{KITTI Road:}
The KITTI Road dataset \citep{kitti} consists of 289 images with manually annotated ground truth labels. 
The setup for the KITTI Road dataset is with the CityScapes.

\textbf{ECSSD:}
ECSSD dataset \citep{ecssd} contains 1,000 complex images with objects of different sizes.
Image labeling is two classes, dividing into foreground and background.

\subsubsection{Performance Metrics}

Mean Intersection-over-Union (mIOU) is the ratio of the intersection and union of two sets of true and predicted values.
\begin{equation}
mIOU=\frac{1}{k+1} \sum_{i=0}^{k} \frac{p_{i i}}{\sum_{j=0}^{k} p_{i j}+\sum_{j=0}^{k} p_{ji}-p_{i i}}
\end{equation}
where $k$ is the category number, $i$ is the true label, and $j$ is the predicted value.
$p_{i j}$ denotes the number of pixels of category $i$ predicted to category $j$.

maxF is a commonly used evaluation metric in salient object detection, which uses dynamic thresholds between 0 and 1, and is generally divided by 255 uniform thresholds.

The normalized compute complexity (NCC) is defined as the optimal utilization of XNOR operations in a compute unit.   
Parameters and computational complexity are derived from theoretical calculations as in the method section.

\subsection{Experimental Results}
\begin{table}[H]
\centering
  \caption{Performance on the driveable area segmentation.}
  \label{table_1}
\scalebox{0.91}
{
\small
  \begin{threeparttable}
   \begin{tabular}{c|c|c|c|c}
    \toprule
  Dataset&Model &  Parameters [MB]   &NCC [$\times 10^{9}$]& mIOU [\%] \\
  \midrule
  \midrule
  \multirow{8}{*}{Cityscapes}  &FCN-8s \citep{fcn}&22.60&10.47&96.94\\
  &Deeplabv3 \citep{deeplabv3}&14.63&14.48&97.30\\
  &PSPNet \citep{pspnet}&12.33&12.97&97.12\\
  &U-Net \citep{unet}&23.26&11.79&97.50\\
    \cline{2-5}
  &FCN8s-XNOR &1.41&0.66&93.10\\
  &Binary DAD-Net \citep{frickenstein2020binary}&1.22&0.73&93.60\\
  &Fixed-Point U-Net \citep{unetfix}&2.31&0.76&91.03\\
  &BNN &0.56&0.36&79.80\\
  &TernaryNet&1.15&0.97&84.80\\
  &gruopNet \citep{group}&1.27&0.77&94.80\\
  & BiFSMNv2\citep{qin2023bifsmnv2}&1.47&1.27&95.32\\
  & Distribution\citep{qin2023distribution}&2.92&3.41&95.49\\
  & EBSR\citep{wei2023ebsr}&1.44&2.59&96.11\\
  &ours&1.32&0.79&\textbf{97.41}\\
    \midrule
     \midrule
  \multirow{8}{*}{KITTI}&FCN-8s \citep{fcn}&22.60&5.250&95.43\\
  &Deeplabv3 \citep{deeplabv3}&14.63&10.46&94.45\\
  &PSPNet \citep{pspnet}&12.33&12.97&95.12\\
  &U-Net \citep{unet}&23.26&5.36&93.26\\
  \cline{2-5}
  &FCN8s-XNOR &1.41&0.33&92.10\\
  &Binary DAD-Net \citep{frickenstein2020binary}&1.22&0.53&92.60\\
  &Fixed-Point U-Net \citep{unetfix}&2.31&0.76&88.10\\
  &BNN&0.56&0.26&74.36\\
  &TernaryNet \citep{3dq}&1.15&0.55&81.70\\
  &gruopNet \citep{group}&1.27&0.38&92.41\\
  & BiFSMNv2\citep{qin2023bifsmnv2}&1.47&0.46&93.11\\
  & Distribution\citep{qin2023distribution}&2.92&1.55&94.34\\
  & EBSR\citep{wei2023ebsr}&1.44&1.27&95.24\\
  &ours&1.32&0.39&\textbf{96.34}\\
\midrule
\bottomrule
\end{tabular}
 \end{threeparttable}
}
\end{table}

\begin{table}[H]
\centering
  \caption{Performance on salient object detection.}
  \label{table_3}
\scalebox{0.91}{
  \small
  \begin{threeparttable}
   \begin{tabular}{c|c|c|c|c}
    \toprule
 Dataset &Model &  Parameters [MB] & NCC [$\times 10^{9}$]& MaxF [$\uparrow$]\\
  \midrule
  \midrule
  \multirow{8}{*}{ECSSD}&FCN-8s \citep{fcn}&22.60&10.47&0.823\\
  &Deeplabv3 \citep{deeplabv3}&14.63&14.48&0.847\\
 &PSPNet \citep{pspnet}&12.33&12.97&0.867\\
  &U-Net \citep{unet}&23.26&11.79&0.834\\
  \cline{2-5}
  &FCN8s-XNOR&1.41&0.66&0.782\\
  &Binary DAD-Net \citep{frickenstein2020binary}&1.22&0.73&0.822\\
  &Fixed-Point U-Net \citep{unetfix}&2.31&0.76&0.786\\
  &BNN&0.56&0.36&0.744\\
  &TernaryNet \citep{3dq}&1.15&0.73&0.793\\
 &gruopNet \citep{group}&1.27&0.48&0.831\\
  & BiFSMNv2\citep{qin2023bifsmnv2}&1.47&0.65&0.844\\
  & Distribution\citep{qin2023distribution}&2.92&1.93&0.847\\
  & EBSR\citep{wei2023ebsr}&1.44&1.41&0.847\\
 &ours&1.32&0.50&\textbf{0.853}\\
\midrule
\bottomrule
\end{tabular}
 \end{threeparttable}
}
\end{table}
The ambition of quantization is to reduce the storage of the model and speed up the inference while ensuring the accuracy.
As for the inference speed, we use the normalized compute complexity (NCC) to present it.
For the storage, a full-precision network storage one number generally cost $32$ bits, while a binary network only needs $1$ bit, so the storage size is $K/32$ of the full precision network, $K$ is the number of branches. Storage is shown in parameter (MB).

We first introduce the results of our proposed upsampling, then introduce the results with the proposed attention, and finally analyze the bottlenecks affecting the accuracy of the binarized semantic segmentation network.
The experimental results are shown in Table \ref{table_1} and Table \ref{table_3}. It can be seen that the proposed binarization network is highly competitive with the full-precision network in terms of result accuracy and also has advantages with other quantization networks in terms of network compression and acceleration.

Drivable area detection in Cityscape and KITTI is focused on autonomous driving scenarios. The drivable area is a two-class task, the purpose of which is to divide pixels into drivable areas and non-drivable areas. The results are shown in Table \ref{table_1}, where Deeplabv3, PSPNet, and U-Net are full precision networks.
FCN8s-XNOR, Binary DAD-net, fixed-point U-net, and BNN are the previous quantization networks. 
FCN8s-XNOR is based on FCN, and Binary DAD-net is based on Deeplabv3. Fixed-point U-net, and BNN are based on U-net.
BiFSMNv2\citep{qin2023bifsmnv2}, Distribution\citep{qin2023distribution} and EBSR \cite{wei2023ebsr} are the representative methods in recent years.
The size of the model parameters of the full precision network is between 12MB-22MB, the accuracy is above 95\%, and the computational complexity is above $10 \times 10^{9}$.
The quantized network model is $\frac{1}{10}$ of the full accuracy network, and the computational complexity is $\frac{1}{20}$, but the accuracy is greatly reduced from the original 97\% to 90\%. Our work not only maintains the accuracy of the full-precision network but also obtains the excellent speed and storage of the quantization network.

Salient object detection is mainly used on mobile devices, and the results are shown in Table \ref{table_3}. It can be seen that the performance is similar to that of the driving area segmentation, which our approach is also very obvious and remarkable.
This shows that our approach is not limited to the autonomous driving field but can significantly improve any intensive output task that requires upsampling.



\subsection{Ablation Study}

The two main contributions of our work are specially designed upsampling and attention, so the ablation study is to demonstrate the significance of both, respectively.
We verify the meaning of upsampling firstly and then attention.

\subsubsection{The effects of upsampling}

\begin{table}[!h]
\centering
  \caption{Performance on the Cityscapes driveable area set.}
  \label{table_up}
   \begin{tabular}{c|c|c}
    \toprule
  Upsample method & mIOU [\%]&$\Delta$ [\%]\\
  \midrule
 Binary &93.33&-\\
   Full-precision&94.70&+1.37\\
   Ours&97.41&+4.08\\
\midrule
\bottomrule
\end{tabular}
\end{table}
To demonstrate the validity of the designed upsampling method, we compare them with the full-precision upsampling, the common binary quantization network, respectively.
Full precision upsampling means that each branch is upsampled and then added.
But the sum is not binary but full precision.
Then the full precision results are quantized, and the quantized results are input into the next stage.
As shown in \ref{table_up}, our proposed upsampling method also has advantages over the full precision upsampling method.
The proposed upsampling can greatly reduce the accuracy loss caused by binarization; thus the results can be competitive with the full precision results.

\subsubsection{The effect of attention}
\begin{table}[!h]
\centering
  \caption{Performance on the Cityscapes.}
  \label{table_attention}
   \begin{tabular}{c|c|c}
    \toprule
  Attention & mIOU [\%]&$\Delta$ [\%]\\
  \midrule
 without attention &94.70&-\\
   single binary attention&94.88&+0.18\\
   our binary attention&97.41&+2.71\\
\midrule
\bottomrule
\end{tabular}
\end{table}

The results are shown in Table \ref{table_attention}.
Since the binary attention mechanism is high-speed and low-power consuming but has a limited degree of differentiation, we calculate attention at all feature levels. Through the above analysis, the calculation speed of binary attention is hundreds of fold fast of full precision, so the inference time and computational complexity are only a little increased although in various levels.
It can be seen that the benefit of a common single-branch attention structure is small, and our proposed binary attention can obtain a 0.53\% improvement, which is similar to the improvement brought by the attention of full-precision.
The results demonstrate the effectiveness of our binary attention with one calculation and multi-distribution strategy.

\subsection{Bottleneck analysis}
\begin{table}[!h]
\centering
  \caption{Performance on Cityscapes driveable area set.}
  \label{table_bottle}   
  \small
   \begin{tabular}{c|c|c|c|c|c}
    \toprule
    \multirow{2}*{Model}&\multicolumn{2}{|c|}{Binary}& \multirow{2}*{mIOU [\%]} & \multirow{2}*{NCC [$ \times 10^{9}$]}&\multirow{2}*{Mem [MB]}\\
   &  Enc&Dec& && \\
     \midrule
ResNet18+DeepLab&\XSolid&\XSolid&97.30&14.48&14.63\\
  \midrule
ResNet18+DeepLab&\Checkmark&\XSolid&96.43&4.83&7.30\\
  \midrule
ResNet18+DeepLab&\XSolid&\Checkmark&96.93&3.46&6.73\\
    \midrule
ResNet18+DeepLab&\Checkmark&\Checkmark&93.26&0.65&1.15\\
    \midrule
  Binary DAD-Net &\Checkmark&\Checkmark &93.60&0.73&1.22\\
    \midrule
  ours &\Checkmark&\Checkmark&97.41&0.79&1.32\\
\midrule
\bottomrule
\end{tabular}
\end{table}

The semantic segmentation networks can be divided into backbones and decode heads.
In this section, we analyze which part or which operation is the bottleneck to limit the accuracy.
The results are shown in Table \ref{table_bottle}, and the following conclusions can be drawn from the experimental results: 1) The chief factor affecting the accuracy is the binarization of the backbone, and the proposed multi-branch parallel method is not the bottleneck, at least for now. 2) The structure of ResNet is not suitable for BNN. 
This phenomenon is the same as the conclusion in \citep{singh2020learning, bethge2018learning}, but the decrease in accuracy of segmentation is more obvious than in the classification.

\section{Conclusion}
Aiming at the characteristics of dense prediction tasks, we designed a multi-branch parallel upsampling method, which retains the original information to the greatest extent, and does not lose the powerful and excellent compression and acceleration of the binary network.
Compared with the full-precision network and multi-valued quantization network, the proposed structure has excellent advantages in results accuracy, model compression, and acceleration.
At the same time, we transform the attention mechanism that is widely used in dense prediction tasks from full precision network to BNN and keep the original great effect, which is of great reference value to other transforms.
We analyze the bottleneck of segmentation accuracy of BNN and draw a conclusion that designing a specialized binary network for semantic segmentation is a valuable direction.
Our work can be seen as a benchmark for BNN in semantic segmentation.

\bibliography{main}

\end{document}